\documentclass[manuscript]{acmart}
\usepackage{todonotes}

\newcommand{\secref}[1]{\mbox{Section~\ref{#1}}}

\newcommand{\tabref}[1]{\mbox{Table~\ref{#1}}}
\newcommand{\figref}[1]{\mbox{Figure~\ref{#1}}}

\AtBeginDocument{%
  \providecommand\BibTeX{{%
    \normalfont B\kern-0.5em{\scshape i\kern-0.25em b}\kern-0.8em\TeX}}}

\copyrightyear{2020}
\acmYear{2020} 
\setcopyright{acmcopyright}
\acmConference[RecSys Challenge '20 Companion]{Companion Proceedings of the 2020 RecSys Conference}{September 22--26, 2020}{Online}
\acmBooktitle{RecSys Challenge '20 Companion: Companion Proceedings of the 2020 RecSys Conference, September 22--26, 2020, Online}
\acmPrice{}
\acmDOI{}
\acmISBN{}

\begin{document}

\title{Two Stages Approach for Tweet Engagement Prediction}

\author{Amine Dadoun}
\affiliation{%
  \institution{EURECOM}
  \city{Sophia Antipolis}
  \country{France}}
\affiliation{%
  \institution{Amadeus SAS}
  \city{Biot}
  \country{France}}
\email{amine.dadoun@eurecom.fr}

\author{Ismail Harrando}
\affiliation{%
  \institution{EURECOM}
  \city{Sophia Antipolis}
  \country{France}}
\email{ismail.harrando@eurecom.fr}

\author{Pasquale Lisena}
\affiliation{%
  \institution{EURECOM}
  \city{Sophia Antipolis}
  \country{France}}
\email{pasquale.lisena@eurecom.fr}

\author{Alison Reboud}
\affiliation{%
  \institution{EURECOM}
  \city{Sophia Antipolis}
  \country{France}}
\email{alison.reboud@eurecom.fr}

\author{Rapha{\"{e}}l Troncy}
\affiliation{%
  \institution{EURECOM}
  \city{Sophia Antipolis}
  \country{France}}
\email{raphael.troncy@eurecom.fr}

\renewcommand{\shortauthors}{Dadoun et al.}

\begin{abstract}
This paper describes the approach proposed by the D2KLab team for the 2020 RecSys Challenge on the task of predicting user engagement facing tweets. This approach relies on two distinct stages. First, relevant features are learned from the challenge dataset. These features are heterogeneous and are the results of different learning modules such as handcrafted features, knowledge graph embeddings, sentiment analysis features and BERT word embeddings. Second, these features are provided in input to an ensemble system based on XGBoost. This approach, only trained on a subset of the entire challenge dataset, ranked 22 in the final leaderboard.
\end{abstract}



\maketitle

\section{Introduction}
\label{sec:introduction}
Dealing with a constantly increasing quantity of information is one of the challenge of modern computer science. The growing amount of content posted on social networks requires the introduction of algorithms that provide end-users with the most relevant content in order to improve their experience. Predicting if a given user would actively interact with a post is a key goal for optimising these algorithms that aim to sustain engagement of the user on the platform.

This paper describes our approach for the task of engagement prediction for the 2020 RecSys Challenge~\cite{recsys2020challenge}. The target dataset -- released in the context of the challenge~\cite{belli2020privacypreserving}  -- includes 160M\footnote{It is worth to note that more than 10\% of the data has been deleted during the course of the challenge and was not processed} public engagements from the Twitter timeline, including both positive (like, retweet, reply, retweet with comment) and negative (absence) examples of engagements. Our method can be described as a two stages approach:
\begin{itemize}
    \item In the first stage, different learning modules extract heterogeneous features from the dataset. Those modules are: handcrafted features extractor, knowledge graph embedding, sentiment analysis and engagement predictions based on tweet content as represented by BERT tokens;
    \item In the second stage, these features are combined in input to an ensemble system, implemented using XGBoost~\cite{chen2016xgboost}.
\end{itemize}
The implementation of this approach is publicly available at \url{https://gitlab.eurecom.fr/dadoun/RecSys_Challenge_2020}.

The remainder of this paper is organised as follows. \secref{sec:approach} presents our approach, while \secref{sec:experiments} details its application to the challenge dataset, together with an internal evaluation protocol and the obtained results. Finally, \secref{sec:conclusion} outlines some conclusions and future work.


\section{Approach}
\label{sec:approach}
\begin{figure}[ht]
    \centering
    \includegraphics[width=\textwidth]{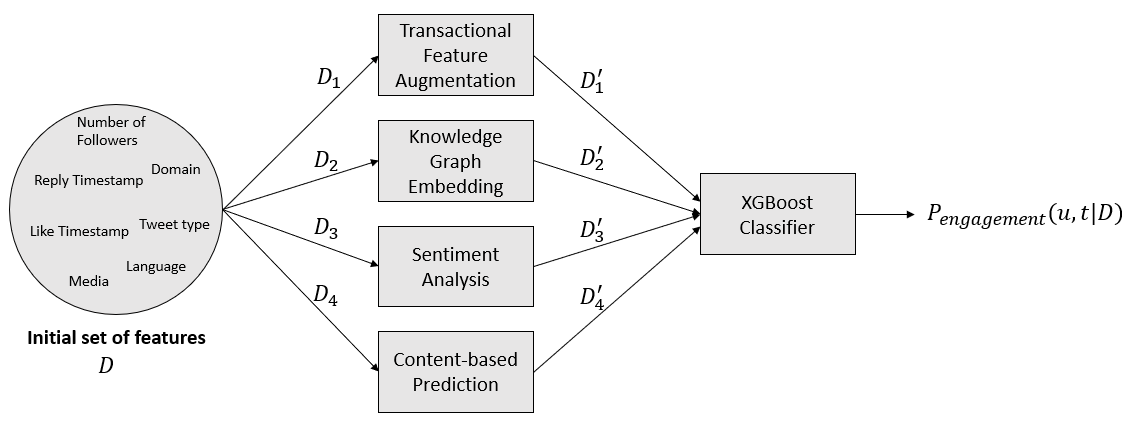}
    \caption{Two stages approach for tweet engagement prediction}
    \label{fig:complete_schema}
\end{figure}

The approach that we propose for predicting the engagement on tweets relies on two subsequent stages, shown in~\figref{fig:complete_schema}. In the first stage, from the set of features $D$ contained in the challenge dataset, we select 4 subsets $D_{i}$. Each of those is processed by a different learning module $i$, which gives in output the set of features $D'_i$. The four modules are detailed respectively in~\secref{subsec:transfeat}, \ref{subsec:kg_embs}, \ref{subsec:sentiment} and \ref{subsec:BERT}. The second stage implements the engagement prediction task. An XGBoost classifier~\cite{chen2016xgboost} is trained on the previously generated features $D_{i}^{'}$ acting as an ensemble classifier. It returns in output the probability that the user $u$ performs an engagement for a tweet (like, retweet, reply, retweet with comment) $P_{engagement}(u,t|D)$. This stage is detailed in~\secref{subsec:ensemble}.

\subsection{Transactional Feature Augmentation}
\label{subsec:transfeat}
In addition to the list of transactional (interaction between users and tweets) features\footnote{This list of transactional features includes: \texttt{present\_domains}, \texttt{tweet\_type}, \texttt{language}, \texttt{present\_media}, \texttt{engagee\_follows\_engager}, \texttt{hashtags}, \texttt{engaging\_user\_follower\_count}, \texttt{engaging\_user\_following\_count}, \texttt{engaged\_with\_user\_follower\_count}, \texttt{engaged\_with\_user\_following\_count}, \texttt{engaging\_user\_account\_creation}, \texttt{engaged\_with\_user\_account\_creation}, \texttt{engaged\_with\_user\_is\_verified}, \texttt{engaging\_user\_is\_verified}.} provided in the released dataset~\cite{belli2020privacypreserving}, we compute some additional features (feature augmentation) that contains more information about the tweets and the transaction (user, tweet). For the sake of explanation, we distinguish between two kinds of users, the \emph{reader} which represent the \texttt{engaged\_with\_user} and the \emph{author} with represent the \texttt{engaging\_user}. We detail below the list of added features: 
\begin{itemize}
    \item Number of engagements per reader: For each engagement (e.g. like, retweet, etc.), we compute the number of times this specific engagement has been performed by the reader before encountering the current tweet.
    \item Number of engagements per author: For each engagement (e.g. like, retweet, etc.), we compute the number of times this specific engagement has been received by the author before posting the current tweet.
    \item Number of engagements per user towards an author: This feature represents the number of times a reader has made an engagement to an author in the past (before seeing the current tweet).
\end{itemize}

\subsection{Knowledge Graph Embedding}
\label{subsec:kg_embs}
Several connections can be seen in the challenge dataset: the relationship \textit{follower-followed} between users, the authorship of a tweet, the interaction with another one, the sharing of hashtags or domains among tweets. Knowledge graphs (KG) provide a suitable way for representing these connections and they have already largely been used to model social networks~\cite{upadhyay2016socialgraph,fafalios2018tweetskb}. KGs have also successfully been exploited in recommender systems, in particular using graph-embeddings~\cite{monti2018recsyschallenge,palumbo2020entity2rec,liu2019graphembeddings}.

\begin{figure}[ht]
    \centering
    \includegraphics[width=0.9\textwidth]{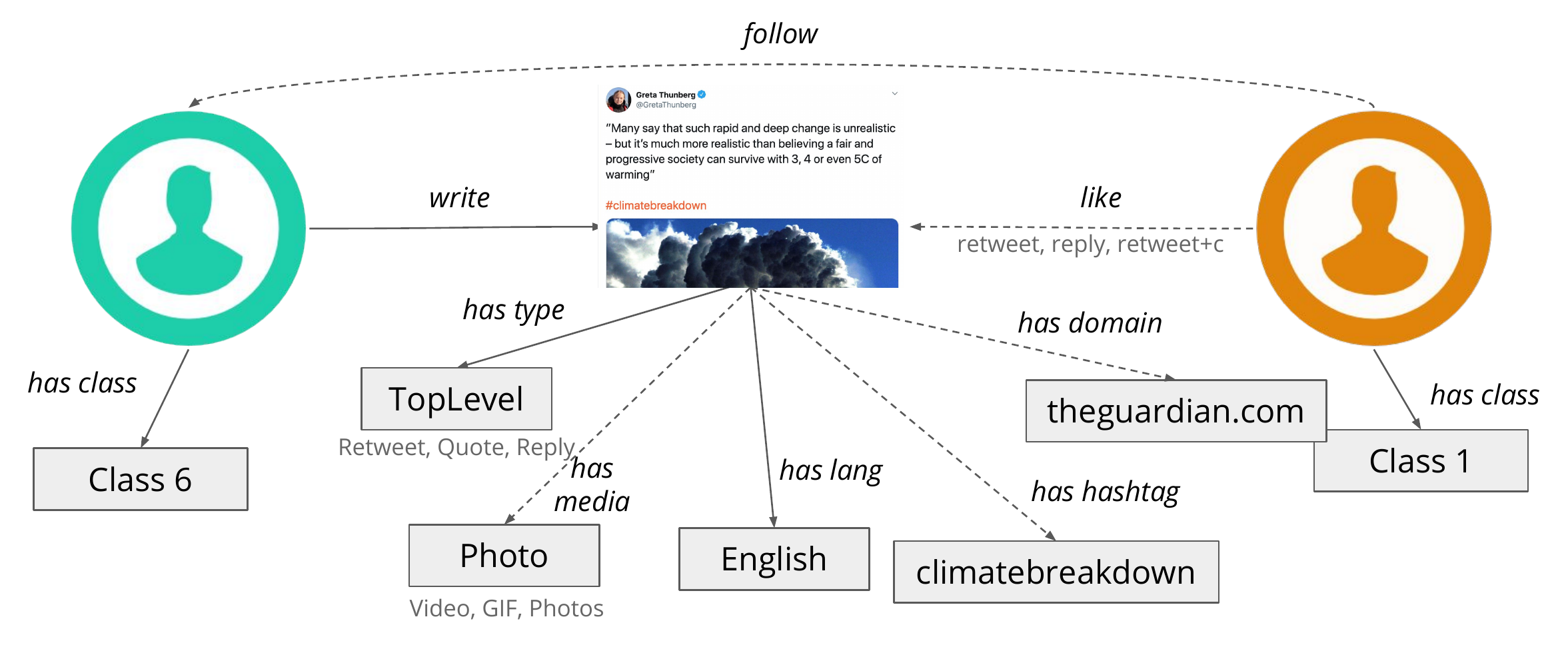}
    \caption{Excerpt of the knowledge graph. Most of the values are de-anonymised for simplicity, while they are in reality identified with an alphanumeric code (i.e. domain, language).}
    \label{fig:graph}
\end{figure}

We used the information coming from the dataset for populating a KG, whose structure is illustrated in~\figref{fig:graph}. The core of this structure is made of the tweet and of the user. The latter can be, in different moment, either the author of a tweet -- to which it is linked through a \texttt{write} edge -- or the one that reads a tweet -- eventually linked to it through an interaction edge, such as a \texttt{like}. When a user follows another user, a specific edge links the two. In addition, a class is assigned to each user depending on her or his number of followers following the distribution presented in~\tabref{tab:user_classes}. Apart from the edges connecting users, a tweet node is also linked with five literal nodes:
\begin{itemize}
    \item \texttt{has type}: TopLevel, Quote, Retweet, or Reply (the value corresponds to the \texttt{tweet type} column in the dataset);
    \item \texttt{has media}: Photo, Video, GIF, or Photos (when there is more than 1 photo);
    \item \texttt{has lang}: language identifier;
    \item \texttt{has hashtag}: hashtag identifier;
    \item \texttt{has domain}: domain identifier.
\end{itemize}

\begin{table}[ht]
\begin{tabular}{|r|r|r|r|}
\hline
\textbf{CLASS} & \textbf{MAX FOLLOWERS} & \textbf{CLASS} & \textbf{MAX FOLLOWERS} \\ \hline
0 & 150 & 4 & 100,000 \\ \hline
1 & 500 & 5 & 1,000,000 \\ \hline
2 & 1,000 & 6 & 10,000,000 \\ \hline
3 & 10,000 & 7 & 200,000,000 \\ \hline
\end{tabular}
\caption{Classification of users depending on their number of followers}
\label{tab:user_classes}
\end{table}

The KG is populated reading the dataset tsv file line by line and creating node and edge instances when required. For example, the \textit{has domain} link would be present only if the tweet contains a domain link. As a consequence, not all edges are created for each row. \figref{fig:graph} represents always-present edges with a continuous arrow, while dashed arrows mark optional edges.

For being used in input to machine learning algorithms, the graph embedding process transforms the graph structure in a set of multi-dimensional vectors. For this purpose, we used \textit{node2vec}~\cite{grover2016node2vec}, a state-of-the-art algorithm that generates random walks between the nodes of the graph, on which it computes the transition probabilities between nodes, which are mapped into the vector space. In other words, nodes sharing more connections are more likely to be part of the same random walk and consequently are more likely to be close in the computed embedding space. We assigned to each kind of edge a different weight, which impacts on the possibility of its nodes to appear in the same random walk.

The limitations of this approach are: the required resources since the machine needs to load the entire graph in memory, and the long computation times, which grows non linearly with the number of nodes and edges. In order to obtain results in a reasonable time for the challenge, we performed the training of these embeddings only on a subset of 40M dataset entries, taking into account only some kind of edges, namely \texttt{follow}, \texttt{write}, \texttt{like}, \texttt{has domain}, and \texttt{has hashtag}.

\subsection{Sentiment Analysis}
\label{subsec:sentiment}
The task of sentiment analysis aims at attributing a predefined sentiment category to a text sequence. In our particular case, it means assigning a positive or negative polarity to each tweet. BERT is very effective for text classification. However, the available pre-trained BERT models for sentiment analysis have been trained on general corpora such as the BooksCorpus (800M words)~\cite{zhu2015aligning} and English Wikipedia (2,500M words), while a fine-tuned model on in-domain data may increase its efficiency. We searched for a domain dataset, containing annotations for the task of sentiment analysis, with two main requirements: \emph{i)} it must contain tweets as they represent a specific type of expression, written in a certain style and with a length constraint and \emph{ii)} the dataset must contain more than 30 languages.

If we have been able to find a sentiment analysis dataset for English tweets called Sent140~\cite{go2009twitter}, to the best of our knowledge, a sentiment analysis dataset matching the language distribution of the challenge dataset does not exist. Given that English still represents approximately 40\% of the tweets in the dataset, we used the Sent140 dataset in the prediction of sentiment labels for those tweets, ignoring those written in other languages.

In~\cite{sun2019fine}, a fine-tuning method for pre-trained BERT models achieves state of the art results for a variety of tasks and datasets. This approach consists in two steps: 
\begin{enumerate}
    \item further training BERT on within-task training data, in our case the tweets;
    \item fine-tuning BERT for the target task using labels, in our case sentiment polarity.
\end{enumerate}
The authors use this method for fine-tuning a pre-traind BERT, providing as input a dataset of film reviews from the IMDB Dataset~\cite{maas2011learning}, annotated with the sentiment polarity. The result is a BERT model specialised for sentiment analysis. 
We performed a further fine-tuning to this model on the Sent140 dataset, in order to capture Twitter-specific expressions and style. The final model has been used on the BERT tokens from the challenge dataset, once decoded to plain text. We used both the sentiment analysis labels predictions and corresponding logits as predictive features for the ensemble.

\subsection{Content-based prediction from BERT Tokens}
\label{subsec:BERT}
The content of tweets was provided in the dataset as a list of BERT~\cite{devlin-etal-2019-bert} token IDs corresponding to multilingual words (e.g. \(``token 21601" \rightarrow  "spiel"\)) or sub-words (\("token 21603" \rightarrow  ``-sation"\)). These tokens can be decoded into strings using the appropriate BERT Tokenizer (\texttt{bert-base-multilingual-cased}) or alternatively used as-is to represent the tweet content. In our work, we implemented two distinct methods for exploiting these tokens:
\begin{itemize}
    \item We fed them into a pre-trained multilingual BERT to generate a fixed-length representation of the tweet textual content, either by pooling (e.g. averaging) the transformed token representations at the output of the BERT model, or by taking the \texttt{[CLS]} embeddings\footnote{When given a sentence as input, the BERT model outputs a contextual embedding for each token of that sentence, as well as a ``sentence-wide'' representation for classification purposes (represented by the special token [CLS]). According to both the original paper and our experiments, both representations generate comparable results when fed to a model to predict user engagement.} which somehow represents the entire input sequence. Both representations are dense 768-dimensional vectors.
    \item We apply the list of tokens as a bag of tokens in input to a TF-IDF model, which uses the count of each token in the tweet and normalises it by the token count in the entire dataset. Since decoding the tokens into their original form increases the vocabulary significantly, we opted for directly using the tokens as represented by their IDs. We also keep the highly-occurring token n-grams ($n <= 3$). This generates a 1M-dimensional sparse tweet representation.
\end{itemize}

Both fixed-size representations are then fed into models to predict the interaction with the tweet (one classifier for each interaction, with a binary output). We use a SVM classifier with the (sparse) TF-IDF features and a feed-forward neural network with the BERT embeddings. The output of these models have been used as a feature ($D'_4$) into the ensemble system.

\subsection{XGBoost}\label{subsec:ensemble}
XGBoost~\cite{chen2016xgboost} is an implementation of gradient boosting decision trees. As presented in~\figref{fig:complete_schema}, XGBoost takes as input the outputs of the first stage modules $D_{i}^{'}$. We detail below the different outputs of stage 1:
\begin{itemize}
    \item $D_{1}^{'}$: The output of \textbf{transactional feature augmentation} module, represented by features coming from the challenge dataset and from extracted transactional features (\secref{subsec:transfeat});
    \item $D_{2}^{'}$: The \textbf{Knowledge Graph Embedding} computed in \secref{subsec:kg_embs} representing readers, authors and tweets;
    \item $D_{3}^{'}$: The outputs of the \textbf{sentiment analysis} module are labels predictions and corresponding logits (\secref{subsec:sentiment});
    \item $D_{4}^{'}$:The softmax score of the \textbf{SVM classifier} trained on the TF-IDF model representing the text tokens (\secref{subsec:BERT}).
\end{itemize}

For each engagement prediction task, an ablation study is performed for features selection in order to train the model on a subset of features $D_{engagement} \subset (D_{1}^{'} \cup D_{2}^{'}  \cup D_{3}^{'} \cup D_{4}^{'})$. This helps speed up the training and also improve slightly PRAUC \& RCE scores. Moreover, we performed a grid-search to find optimal hyper-parameters of XGBoost classifier.

\section{Experiments}
\label{sec:experiments}
In this section, we discuss how we have implemented our approach and we comment on the obtained results. Following the challenge rules, the evaluation relies on two metrics: the area under precision-recall curve (PRAUC) and the relative cross-entropy (RCE)~\cite{belli2020privacypreserving}.

\subsection{Development Pipeline}\label{subsec:dev_pipeline}
\begin{figure}[ht]
    \centering
    \includegraphics[width=\textwidth]{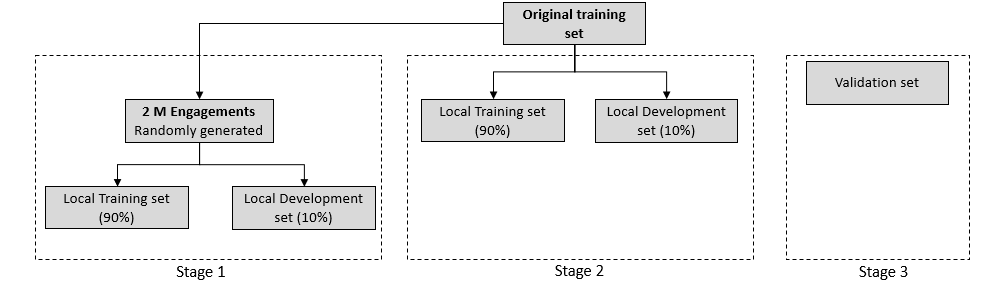}
    \caption{Development Pipeline}
    \label{fig:Evaluation_protocol}
\end{figure}

The challenge dataset consists of a training (121M public engagements) and validation set (12M), together with a final submission set (12M) released in the last part of the challenge. Only the training set contains the information about the engagements. To enable a faster computation and evaluation of planned experiments, we relied on an development pipeline composed of three stages represented in \figref{fig:Evaluation_protocol}. In each stage, training and evaluation are performed in different subset of the original training set:
\begin{enumerate}
    \item In the first stage, we use a randomly extracted subset from the training set that including 2 million rows (i.e. public engagements) and split it into a training set (90\%) and a development set (10\%). This phase allowed to perform experiments with different methods and feature sets.
    \item When an improvement was observed in both PRAUC and RCE, we moved to a second stage where the full original training set was used, split again into local training set (90\%) and local development set (10\%). This phase helped us to understand if the method developed during the first stage is generalizable to a larger dataset.
    \item Finally, in the last stage, we use the trained models to compute predictions on the original validation set. These results are submitted to the (public) challenge leaderboard.
\end{enumerate}

Because of time and hardware constraints, we used in Stage 2 a local training set corresponding to only 40\% (instead of 90\%) of the original dataset. This additional sampling has enabled us to study the results obtained by the KG embedding and content-based prediction from BERT tokens modules. In fact, we could not scale to the entire training set with these modules due to memory limitations.

\subsection{Results and Discussion}\label{subsec:results}
\tabref{tab:results_dev} presents the results obtained on local development dataset from the different implemented models trained in Stage 2:
\begin{itemize}
    \item Model 1: XGBoost trained on only $D_{1}^{'}$.
    \item Model 2: XGBoost trained on $D_{1}^{'}$ \& $D_{2}^{'}$.
    \item Model 3: XGBoost trained on $D_{1}^{'}$ \& $D_{3}^{'}$. 
    \item Model 4: XGBoost trained on $D_{1}^{'}$ \& $D_{4}^{'}$. 
\end{itemize}

\begin{table}[htbp]
\begin{tabular}{|r|r|r|r|r|r|r|r|r|}
\hline
\textbf{Model} & \shortstack{\textbf{PRAUC} \\ \textbf{Retweet}} & \shortstack{\textbf{RCE} \\ \textbf{Retweet}} & \shortstack{\textbf{PRAUC} \\ \textbf{Reply}} & \shortstack{\textbf{RCE} \\ \textbf{Reply}} & \shortstack{\textbf{PRAUC} \\ \textbf{Like}} & \shortstack{\textbf{RCE} \\ \textbf{Like}} & \shortstack{\textbf{PRAUC} \\ \textbf{Retweet} \\ \textbf{with Comment}} & \shortstack{\textbf{RCE} \\  \textbf{Retweet} \\ \textbf{with Comment}} \\ \hline
Model 1 & 0.66 & 40.03 & 0.33 & \textbf{26.44} & 0.86 & 38.63 & 0.18 & \textbf{17.71} \\ \hline
Model 2 & 0.16 & -21.35 & 0.22 & -56.22 & 0.43 & -12.84 & 0.05 & -205.19 \\ \hline
Model 3 & 0.64 & 38.65 & 0.27 & 22.14 & 0.80 & 34.23 & 0.16 & 15.20 \\ \hline
Model 4 & \textbf{0.68} & \textbf{42.03} & \textbf{0.33} & 25.92 & \textbf{0.89} & \textbf{41.08} & \textbf{0.19} & 16.51 \\ \hline
\end{tabular}
\caption{Models evaluated on our local development set (10\% of the original training set)}
\label{tab:results_dev}
\end{table}

We only submitted to the public leaderboard the models 1 and 4 which were the most accurate ones. As we can see from the values in \tabref{tab:results_val}, PRAUC and RCE scores were remarkably different. We believe that the reason for this is the different variables distribution between the training and validation sets based on an exploratory data analysis we performed on the datasets.

\begin{table}[htbp]
\begin{tabular}{|r|r|r|r|r|r|r|r|r|}
\hline
\textbf{Model} & \shortstack{\textbf{PRAUC} \\ \textbf{Retweet}} & \shortstack{\textbf{RCE} \\ \textbf{Retweet}} & \shortstack{\textbf{PRAUC} \\ \textbf{Reply}} & \shortstack{\textbf{RCE} \\ \textbf{Reply}} & \shortstack{\textbf{PRAUC} \\ \textbf{Like}} & \shortstack{\textbf{RCE} \\ \textbf{Like}} & \shortstack{\textbf{PRAUC} \\ \textbf{Retweet} \\ \textbf{with Comment}} & \shortstack{\textbf{RCE} \\  \textbf{Retweet} \\ \textbf{with Comment}} \\ \hline
Model 1 & 0.28 & 7.82 & 0.07 & 5.25 & 0.66 & 10.25 & 0.02 & -25.07 \\ \hline
Model 4 & 0.32 & 7.98 & \textbf{0.14} & \textbf{7.82} & 0.66 & 12.21 & 0.02 & -22.35 \\ \hline
Model 1+ & \textbf{0.39} & \textbf{15.94} & 0.12 & 6.45 & \textbf{0.68} & \textbf{12.44} & \textbf{0.03} & \textbf{-16.71} \\ \hline
\end{tabular}
\caption{Models evaluated on the validation set. Model 1+ was trained using the entire local training set.}
\label{tab:results_val}
\end{table}

We finally used our Model 4 for the final submission on the test data. Even if it is not the better performing one, we believe that it best represents our approach which is to combine features coming from different data sources. This approach ranked at position 22 in the final leaderboard\footnote{\url{https://recsys-twitter.com/final_leaderboard/results}} (\tabref{tab:results_final}). In contrast to the top ranked systems, we observe that our method is able to obtain a better balance between RCE and PRAUC for all predictions.

\begin{table}[htbp]
\begin{tabular}{|r|r|r|r|r|r|r|r|r|}
\hline
\textbf{Model} & \shortstack{\textbf{PRAUC} \\ \textbf{Retweet}} & \shortstack{\textbf{RCE} \\ \textbf{Retweet}} & \shortstack{\textbf{PRAUC} \\ \textbf{Reply}} & \shortstack{\textbf{RCE} \\ \textbf{Reply}} & \shortstack{\textbf{PRAUC} \\ \textbf{Like}} & \shortstack{\textbf{RCE} \\ \textbf{Like}} & \shortstack{\textbf{PRAUC} \\ \textbf{Retweet} \\ \textbf{with Comment}} & \shortstack{\textbf{RCE} \\  \textbf{Retweet} \\ \textbf{with Comment}} \\ \hline
Model 4 & 0.2924 &	5.95 &	0.0789 &	0.25 &	0.6145 &	0.37 &	0.0186 &	-11.40 \\ \hline
\end{tabular}
\caption{Results on the final test set}
\label{tab:results_final}
\end{table}

\section{Conclusion and Future Work}
\label{sec:conclusion}
Predicting user engagement facing tweets on a very large dataset is a challenging task, both in terms of leveraging the massive information available at scale and coming up with representative and significant features to feed the different models. In this paper, we have experimented with a broad set of features we extracted and exploited for this task. 

A key takeaway is the importance of leveraging the entire dataset to have a well-performing model, as our more advanced models, while showing very good results on our local training subset, did not manage to perform as well on the public leaderboard. The effort in studying and understanding the data and its distribution according to key factors has proven to be crucial for realising subsets that are representative of the final validation set. This would also stand as a straightforward solution to the scalability limitation of some of the used models. Other improvements include the capture of time variant information (for example the start and stop of a user following action), which can be represented both in the transactional features and in the KG.

\newpage
\bibliographystyle{ACM-Reference-Format}
\bibliography{bibliography}


\begin{thebibliography}{14}


\ifx \showCODEN    \undefined \def \showCODEN     #1{\unskip}     \fi
\ifx \showDOI      \undefined \def \showDOI       #1{#1}\fi
\ifx \showISBNx    \undefined \def \showISBNx     #1{\unskip}     \fi
\ifx \showISBNxiii \undefined \def \showISBNxiii  #1{\unskip}     \fi
\ifx \showISSN     \undefined \def \showISSN      #1{\unskip}     \fi
\ifx \showLCCN     \undefined \def \showLCCN      #1{\unskip}     \fi
\ifx \shownote     \undefined \def \shownote      #1{#1}          \fi
\ifx \showarticletitle \undefined \def \showarticletitle #1{#1}   \fi
\ifx \showURL      \undefined \def \showURL       {\relax}        \fi
\providecommand\bibfield[2]{#2}
\providecommand\bibinfo[2]{#2}
\providecommand\natexlab[1]{#1}
\providecommand\showeprint[2][]{arXiv:#2}

\bibitem[\protect\citeauthoryear{Belli, Ktena, Tejani, Lung-Yut-Fon, Portman,
  Zhu, Xie, Gupta, Bronstein, Delić, Sottocornola, Anelli, Andrade, Smith, and
  Shi}{Belli et~al\mbox{.}}{2020b}]%
        {belli2020privacypreserving}
\bibfield{author}{\bibinfo{person}{Luca Belli}, \bibinfo{person}{Sofia~Ira
  Ktena}, \bibinfo{person}{Alykhan Tejani}, \bibinfo{person}{Alexandre
  Lung-Yut-Fon}, \bibinfo{person}{Frank Portman}, \bibinfo{person}{Xiao Zhu},
  \bibinfo{person}{Yuanpu Xie}, \bibinfo{person}{Akshay Gupta},
  \bibinfo{person}{Michael Bronstein}, \bibinfo{person}{Amra Delić},
  \bibinfo{person}{Gabriele Sottocornola}, \bibinfo{person}{Walter Anelli},
  \bibinfo{person}{Nazareno Andrade}, \bibinfo{person}{Jessie Smith}, {and}
  \bibinfo{person}{Wenzhe Shi}.} \bibinfo{year}{2020}\natexlab{b}.
\newblock \bibinfo{title}{Privacy-Preserving Recommender Systems Challenge on
  Twitter's Home Timeline}.
\newblock
\newblock
\showeprint[arxiv]{2004.13715}


\bibitem[\protect\citeauthoryear{Belli, Ktena, Tejani, Lung-Yut-Fon, Portman,
  Zhu, Xie, Gupta, Bronstein, Delić, Sottocornola, Anelli, Andrade, Smith, and
  Shi}{Belli et~al\mbox{.}}{2020a}]%
        {recsys2020challenge}
\bibfield{author}{\bibinfo{person}{Luca Belli}, \bibinfo{person}{Sofia~Ira
  Ktena}, \bibinfo{person}{Alykhan Tejani}, \bibinfo{person}{Alexandre
  Lung-Yut-Fon}, \bibinfo{person}{Frank Portman}, \bibinfo{person}{Xiao Zhu},
  \bibinfo{person}{Yuanpu Xie}, \bibinfo{person}{Akshay Gupta},
  \bibinfo{person}{Michael Bronstein}, \bibinfo{person}{Amra Delić},
  \bibinfo{person}{Gabriele Sottocornola}, \bibinfo{person}{Walter Anelli},
  \bibinfo{person}{Nazareno Andrade}, \bibinfo{person}{Jessie Smith}, {and}
  \bibinfo{person}{Wenzhe Shi}.} \bibinfo{year}{2020}\natexlab{a}.
\newblock \bibinfo{title}{RecSys Challenge 2020}.
\newblock
\newblock


\bibitem[\protect\citeauthoryear{Chen and Guestrin}{Chen and Guestrin}{2016}]%
        {chen2016xgboost}
\bibfield{author}{\bibinfo{person}{Tianqi Chen} {and} \bibinfo{person}{Carlos
  Guestrin}.} \bibinfo{year}{2016}\natexlab{}.
\newblock \showarticletitle{XGBoost: A Scalable Tree Boosting System}. In
  \bibinfo{booktitle}{\emph{22$^{nd}$ ACM SIGKDD International Conference on
  Knowledge Discovery and Data Mining (KDD)}}. \bibinfo{address}{San Francisco,
  California, USA}, \bibinfo{pages}{785--–794}.
\newblock


\bibitem[\protect\citeauthoryear{Devlin, Chang, Lee, and Toutanova}{Devlin
  et~al\mbox{.}}{2019}]%
        {devlin-etal-2019-bert}
\bibfield{author}{\bibinfo{person}{Jacob Devlin}, \bibinfo{person}{Ming-Wei
  Chang}, \bibinfo{person}{Kenton Lee}, {and} \bibinfo{person}{Kristina
  Toutanova}.} \bibinfo{year}{2019}\natexlab{}.
\newblock \showarticletitle{{BERT}: Pre-training of Deep Bidirectional
  Transformers for Language Understanding}. In
  \bibinfo{booktitle}{\emph{57$^{th}$ Annual Meeting of the Association for
  Computational Linguistics: Human language technologies}}.
  \bibinfo{address}{Minneapolis, Minnesota}, \bibinfo{pages}{4171--4186}.
\newblock


\bibitem[\protect\citeauthoryear{Fafalios, Iosifidis, Ntoutsi, and
  Dietze}{Fafalios et~al\mbox{.}}{2018}]%
        {fafalios2018tweetskb}
\bibfield{author}{\bibinfo{person}{Pavlos Fafalios}, \bibinfo{person}{Vasileios
  Iosifidis}, \bibinfo{person}{Eirini Ntoutsi}, {and} \bibinfo{person}{Stefan
  Dietze}.} \bibinfo{year}{2018}\natexlab{}.
\newblock \showarticletitle{TweetsKB: A Public and Large-Scale RDF Corpus of
  Annotated Tweets}. In \bibinfo{booktitle}{\emph{European Semantic Web
  Conference (ESWC)}}. \bibinfo{address}{Heraklion, Greece},
  \bibinfo{pages}{177--190}.
\newblock


\bibitem[\protect\citeauthoryear{Go, Bhayani, and Huang}{Go
  et~al\mbox{.}}{2009}]%
        {go2009twitter}
\bibfield{author}{\bibinfo{person}{Alec Go}, \bibinfo{person}{Richa Bhayani},
  {and} \bibinfo{person}{Lei Huang}.} \bibinfo{year}{2009}\natexlab{}.
\newblock \bibinfo{booktitle}{\emph{Twitter sentiment classification using
  distant supervision}}.
\newblock \bibinfo{type}{{T}echnical {R}eport}. \bibinfo{institution}{CS224N
  project report, Stanford}.
\newblock


\bibitem[\protect\citeauthoryear{Grover and Leskovec}{Grover and
  Leskovec}{2016}]%
        {grover2016node2vec}
\bibfield{author}{\bibinfo{person}{Aditya Grover} {and} \bibinfo{person}{Jure
  Leskovec}.} \bibinfo{year}{2016}\natexlab{}.
\newblock \showarticletitle{{node2vec: Scalable Feature Learning for
  Networks}}. In \bibinfo{booktitle}{\emph{{22$^{nd}$ ACM SIGKDD International
  Conference on Knowledge Discovery and Data Mining (KDD)}}}.
  \bibinfo{address}{San Francisco, CA, USA}, \bibinfo{pages}{855–--864}.
\newblock


\bibitem[\protect\citeauthoryear{Liu, Li, Yao, and Tang}{Liu
  et~al\mbox{.}}{2019}]%
        {liu2019graphembeddings}
\bibfield{author}{\bibinfo{person}{Chan Liu}, \bibinfo{person}{Lun Li},
  \bibinfo{person}{Xiaolu Yao}, {and} \bibinfo{person}{Lin Tang}.}
  \bibinfo{year}{2019}\natexlab{}.
\newblock \showarticletitle{A Survey of Recommendation Algorithms Based on
  Knowledge Graph Embedding}. In \bibinfo{booktitle}{\emph{IEEE International
  Conference on Computer Science and Educational Informatization (CSEI)}}.
  \bibinfo{address}{Xinxiang, China}, \bibinfo{pages}{168--171}.
\newblock


\bibitem[\protect\citeauthoryear{Maas, Daly, Pham, Huang, Ng, and Potts}{Maas
  et~al\mbox{.}}{2011}]%
        {maas2011learning}
\bibfield{author}{\bibinfo{person}{Andrew Maas}, \bibinfo{person}{Raymond~E
  Daly}, \bibinfo{person}{Peter~T Pham}, \bibinfo{person}{Dan Huang},
  \bibinfo{person}{Andrew~Y Ng}, {and} \bibinfo{person}{Christopher Potts}.}
  \bibinfo{year}{2011}\natexlab{}.
\newblock \showarticletitle{Learning word vectors for sentiment analysis}. In
  \bibinfo{booktitle}{\emph{49$^{th}$ Annual Meeting of the Association for
  Computational Linguistics: Human language technologies}}.
  \bibinfo{address}{Portland, OR, USA}, \bibinfo{pages}{142--150}.
\newblock


\bibitem[\protect\citeauthoryear{Monti, Palumbo, Rizzo, Lisena, Troncy, Fell,
  Cabrio, and Morisio}{Monti et~al\mbox{.}}{2018}]%
        {monti2018recsyschallenge}
\bibfield{author}{\bibinfo{person}{Diego Monti}, \bibinfo{person}{Enrico
  Palumbo}, \bibinfo{person}{Giuseppe Rizzo}, \bibinfo{person}{Pasquale
  Lisena}, \bibinfo{person}{Rapha\"{e}l Troncy}, \bibinfo{person}{Michael
  Fell}, \bibinfo{person}{Elena Cabrio}, {and} \bibinfo{person}{Maurizio
  Morisio}.} \bibinfo{year}{2018}\natexlab{}.
\newblock \showarticletitle{{An Ensemble Approach of Recurrent Neural Networks
  using Pre-Trained Embeddings for Playlist Completion}}. In
  \bibinfo{booktitle}{\emph{12$^{th}$ ACM Conference on Recommender Systems
  (RecSys), Challenge Track}}. \bibinfo{address}{Vancouver, BC, Canada}.
\newblock


\bibitem[\protect\citeauthoryear{Palumbo, Monti, Rizzo, Troncy, and
  Baralis}{Palumbo et~al\mbox{.}}{2020}]%
        {palumbo2020entity2rec}
\bibfield{author}{\bibinfo{person}{Enrico Palumbo}, \bibinfo{person}{Diego
  Monti}, \bibinfo{person}{Giuseppe Rizzo}, \bibinfo{person}{Raphaël Troncy},
  {and} \bibinfo{person}{Elena Baralis}.} \bibinfo{year}{2020}\natexlab{}.
\newblock \showarticletitle{{entity2rec: Property-specific knowledge graph
  embeddings for item recommendation}}.
\newblock \bibinfo{journal}{\emph{Expert Systems with Applications}}
  \bibinfo{volume}{151} (\bibinfo{year}{2020}).
\newblock


\bibitem[\protect\citeauthoryear{Sun, Qiu, Xu, and Huang}{Sun
  et~al\mbox{.}}{2019}]%
        {sun2019fine}
\bibfield{author}{\bibinfo{person}{Chi Sun}, \bibinfo{person}{Xipeng Qiu},
  \bibinfo{person}{Yige Xu}, {and} \bibinfo{person}{Xuanjing Huang}.}
  \bibinfo{year}{2019}\natexlab{}.
\newblock \showarticletitle{{How to fine-tune BERT for text classification?}}.
  In \bibinfo{booktitle}{\emph{18$^{th}$ China National Conference on Chinese
  Computational Linguistics (CCL)}}. \bibinfo{address}{Kunming, China},
  \bibinfo{pages}{194--206}.
\newblock


\bibitem[\protect\citeauthoryear{Upadhyay, Singh, Abhishek, and Singh}{Upadhyay
  et~al\mbox{.}}{2016}]%
        {upadhyay2016socialgraph}
\bibfield{author}{\bibinfo{person}{Shubhnkar Upadhyay},
  \bibinfo{person}{Avadhesh Singh}, \bibinfo{person}{Kumar Abhishek}, {and}
  \bibinfo{person}{M.~P. Singh}.} \bibinfo{year}{2016}\natexlab{}.
\newblock \showarticletitle{{Deploying a Social Web Graph Over a Semantic Web
  Framework}}. In \bibinfo{booktitle}{\emph{{Computational Intelligence in Data
  Mining (CIDM)}}}. \bibinfo{address}{Cap Town, South Africa},
  \bibinfo{pages}{73--83}.
\newblock


\bibitem[\protect\citeauthoryear{Zhu, Kiros, Zemel, Salakhutdinov, Urtasun,
  Torralba, and Fidler}{Zhu et~al\mbox{.}}{2015}]%
        {zhu2015aligning}
\bibfield{author}{\bibinfo{person}{Yukun Zhu}, \bibinfo{person}{Ryan Kiros},
  \bibinfo{person}{Rich Zemel}, \bibinfo{person}{Ruslan Salakhutdinov},
  \bibinfo{person}{Raquel Urtasun}, \bibinfo{person}{Antonio Torralba}, {and}
  \bibinfo{person}{Sanja Fidler}.} \bibinfo{year}{2015}\natexlab{}.
\newblock \showarticletitle{Aligning books and movies: Towards story-like
  visual explanations by watching movies and reading books}. In
  \bibinfo{booktitle}{\emph{IEEE International Conference on Computer Vision
  (ICCV)}}. \bibinfo{pages}{19--27}.
\newblock


\end{thebibliography}

\end{document}